\documentclass{article}

\PassOptionsToPackage{numbers, sort&compress, comma, square}{natbib}
\usepackage[preprint]{neurips_2022}

\usepackage[utf8]{inputenc} 
\usepackage[T1]{fontenc}    
\usepackage{hyperref}       
\usepackage{url}            
\usepackage{booktabs}       
\usepackage{amsfonts}       
\usepackage{nicefrac}       
\usepackage{microtype}      
\usepackage{xcolor}         

\usepackage{amssymb}
\usepackage{latexsym}
\usepackage{mathtools}
\usepackage{algorithm}
\usepackage[noend]{algpseudocode}
\usepackage{setspace}

\usepackage{graphicx}
\usepackage{amsmath}
\usepackage{amsthm}
\usepackage{amssymb}
\usepackage{booktabs}
\usepackage{graphicx} 
\usepackage{graphics}
\usepackage{amssymb}
\usepackage{amsmath}
\usepackage{pifont}
\usepackage{makecell}
\usepackage{resizegather}
\usepackage{etoolbox}
\usepackage{booktabs,makecell,tabularx} 
\usepackage{enumitem}
\usepackage{mathtools}
\usepackage{cleveref}
\usepackage{multirow}
\usepackage{booktabs,array}
\usepackage{amsmath, bm}
\usepackage{color, colortbl}
\usepackage{xcolor}
\usepackage{booktabs}
\usepackage{algorithm}
\usepackage{array}
\usepackage{subfigure}
\usepackage{multirow}
\usepackage{threeparttable}
\usepackage{tabularx}
\usepackage{booktabs}
\usepackage{colortbl}
\usepackage{xcolor}
\usepackage{stfloats}
\usepackage[noend]{algpseudocode}
\usepackage{mathtools}
\usepackage{makecell}
\usepackage{setspace}
\usepackage{resizegather}
\usepackage{etoolbox}
\usepackage{booktabs,makecell,tabularx} 
\usepackage{enumitem}
\usepackage{mathtools}
\usepackage{cleveref}
\usepackage{multirow}
\usepackage{booktabs,array}
\usepackage{amsmath, bm}
\usepackage{bbm}

\usepackage{wrapfig}

\newcommand{\cmark}{\textcolor{green!60!black}{\ding{51}}} 
\newcommand{\xmark}{\textcolor{red!70!}{\ding{55}}} 

\title{Origin Tracing and Detecting of LLMs}

\author{%
  Linyang Li, Pengyu Wang, Ke Ren, Tianxiang Sun, Xipeng Qiu \\
  Department of Computer Science\\
  Fudan University\\
  Shanghai \\
  \texttt{\{linyangli19,xpqiu\}@fudan.edu.cn} \\
}

\begin{document}

\maketitle

\begin{abstract}
  
The extraordinary performance of large language models (LLMs) heightens the importance of detecting whether the context is generated by an AI system.
More importantly, while more and more companies and institutions release their LLMs, the origin can be hard to trace.
Since LLMs are heading towards the time of AGI, similar to the origin tracing in anthropology, it is of great importance to trace the origin of LLMs.
In this paper, we first raise the concern of the origin tracing of LLMs and propose an effective method to trace and detect AI-generated contexts.
We introduce a novel algorithm that leverages the contrastive features between LLMs and extracts model-wise features to trace the text origins.
Our proposed method works under both white-box and black-box settings therefore can be widely generalized to detect various LLMs.(e.g. can be generalized to detect GPT-3 models without the GPT-3 models).
Also, our proposed method requires only limited data compared with the supervised learning methods and can be extended to trace new-coming model origins.
We construct extensive experiments to examine whether we can trace the origins of given texts.
We provide valuable observations based on the experimental results, such as the difficulty level of AI origin tracing, and the AI origin similarities, and call for ethical concerns of LLM providers.
We are releasing all codes and data as a toolkit and benchmark for future AI origin tracing and detecting studies. \footnote{We are releasing all available resource at \url{https://github.com/OpenLMLab/}.} 
\end{abstract}

\section{Introduction}

Using LLMs such as ChatGPT and GPT4 \cite{OpenAI2023GPT4TR} for various daily routines and copilot in work is becoming a new trend that draws worldwide attention not only in the machine learning community.
Starting from GPT \cite{radford2018improving} and BERT \cite{bert}, pre-trained models have developed for several years, and trustworthy and security concerns have been constantly discussed \cite{Bai2022TrainingAH}.
The performances of LLMs are sensational, therefore, the usage of LLMs should be strictly supervised by users as well as service providers. 
One trend in ensuring the safety of LLMs is to build detection tools that can discriminate whether an AI system generates a certain text \cite{Jawahar2020AutomaticDO}.
AI-generated context detection is useful in releasing texts that require strict censoring or originality such as official documents, consultation, and student submissions \cite{Mitchell2023DetectGPTZM} to avoid abuse of AI systems.
Further, a more critical and applicable field is to trace the origin of LLMs.
While more and more companies and institutions are releasing their original LLMs, it is of great importance to trace whether an LLM is trained from a previous model, or is copied or distilled from another LLM.
Since LLMs can produce massive generated data, future LLMs might be trained from these generated data, the human-written texts might be contaminated with different LLMs.
Therefore, tracing the origin of the text is a major challenge in future LLM industries.

In this work, we first introduce the concept of origin tracing. Then we provide an effective tool named \textbf{Sniffer} and its evaluation benchmark to study the origin tracing problem.
\begin{wrapfigure}
{l}{5cm}
{\includegraphics[width=0.35\textwidth]{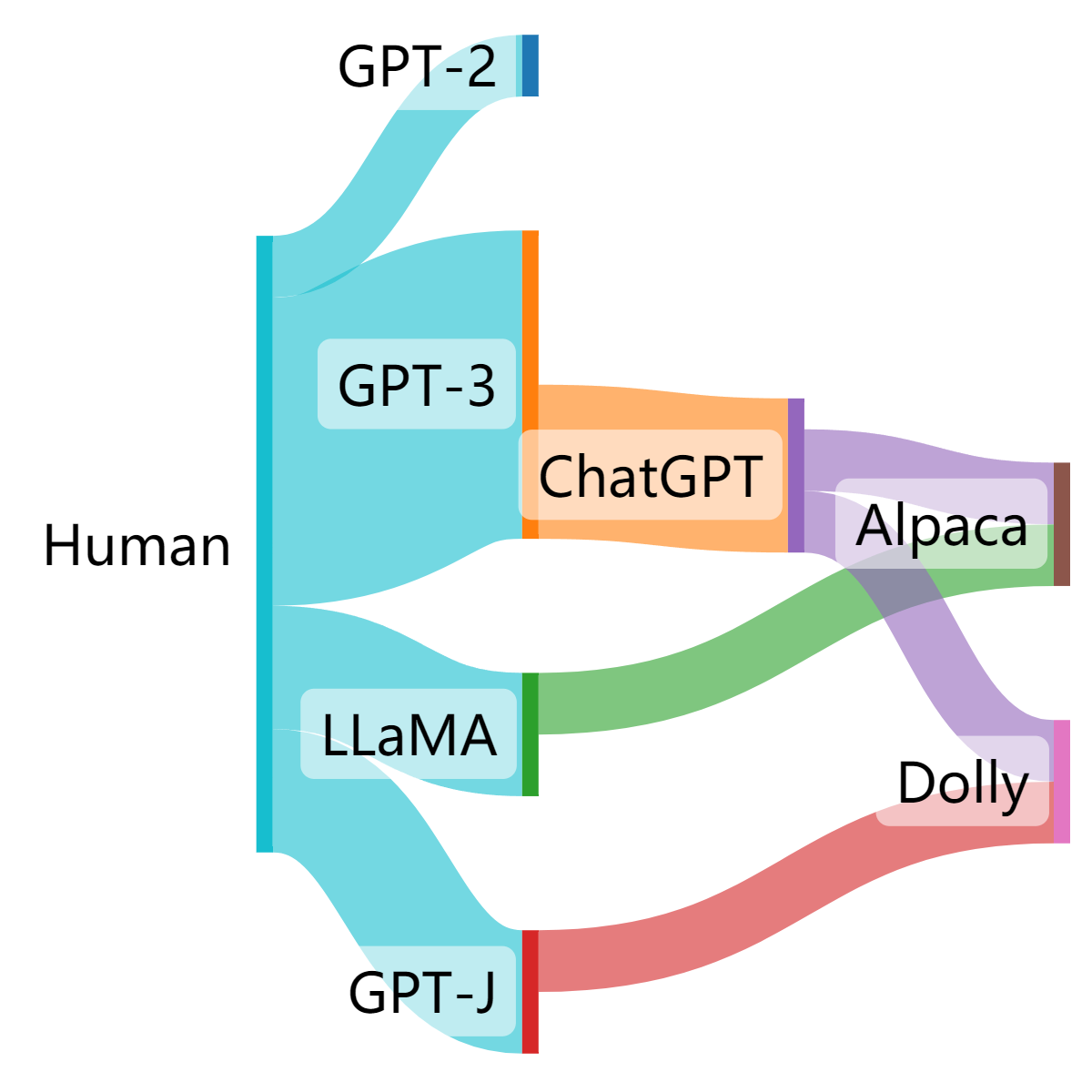}}
\caption{A knowledge flow of LLMs; with origin tracing, we can trace Alpaca back to ChatGPT and LLaMA.}
\vspace{-0.4cm}
\end{wrapfigure}

Origin tracing is to further categorize the origin of a given text.
Specifically, we categorize the origins by the LLM abilities and by LLM service providers.
We first trace the origin of GPT2 level LLMs such as GPT2 \cite{radford2019language} from OpenAI, GPT-Neo/J \cite{Black2021GPTNeoLS,mesh-transformer-jax} from EleutherAI, we can also trace the GPT3 level LLMs such as text-davinci-003, ChatGPT (GPT3.5 turbo) \cite{brown2020language}, LLaMA \cite{Touvron2023LLaMAOA}.
With origin tracing, we can avoid AI abuse or potential model theft since the texts can be traced to a specific service provider. 
While more and more companies and institutions are releasing their LLMs, origin tracing is the keystone of \textit{anthropology} of LLMs. 
Then we introduce \textbf{Sniffer}, the first origin tracing tool. 
In Sniffer, we use contrastive features across open-source LLMs such as GPT2, GPT-Neo/J, and LLaMA.
Specifically, we design heuristic features that capture the model-wise discrepancies which can help trace the origin of given texts.
Then we utilize a simple linear classifier to project the extracted features to specific origins including known origins that have open-source models and unknown origins that are black boxes to users.
The core motivation of the origin tracing tool Sniffer utilizes the discrepancies between LLMs as features to help trace the origins.
Compared with previous methods, Sniffer utilizes model-wise features, which is different from supervised learning methods such as fine-tuning a RoBERTa \cite{liu2019roberta} model;
further, Sniffer is able to trace text origins and generalize to unknown origins, while previous methods that use model-wise features cannot.

With Sniffer, we further introduce a test dataset that contains collected texts from different origins as a benchmark to study the origin tracing problem.  We provide plenty of experiments and through the experimental results, we have several non-trivial observations that help future LLM studies.
In general, we find that: 
(1) we are able to trace the origins of generated texts when we can possess the models;
(2) it grows harder to detect and trace origins when the LLMs are stronger;
(3) LLM providers need to be more cautious as the origin of generated texts might be known only by the providers. 
(4) We can trace the origins of distilled LLMs such such as Alpaca and Dolly.

To summarize, in this paper, we: 
(1) raise the concern of origin tracing of AI-generated contexts; 
(2) build a feature-wise tool to trace the origins of various open-source models by releasing a diversified benchmark for AI-generated contexts detection and origin tracing;
(3) conduct various experiments to analyze the ability of LLMs when they are being traced and hope that future works can pay more attentions to the origin tracing of LLMs.

\vspace{-0.2cm}
\section{Related Work}
\vspace{-0.4cm}
\textbf{Increasing Concern of LLM Security}

 Pre-trained models \cite{bert,radford2018improving,radford2019language,2020t5,OpenAI2023GPT4TR} are supposed to be harmless, harmful, and honest to users \cite{Bai2022TrainingAH}, however, there are various aspects that challenge LLM securities such as social bias, stereotypes, privacy leak or adversarial examples \cite{zhao2017men,Carlini2020ExtractingTD,li2020bert}. 

Detection of AI-generated contexts is also a rapidly growing field that requires attention since the abuse of AI might be a major challenge in LLM applications \cite{Bai2022TrainingAH}.
Still, current methods only consider detection as a binary task, that is, whether a given text is generated by an AI without considering tracing the origins of texts.
Detection methods can be categorized into two lines:

\textbf{Semantic-wise Detection}

The most straightforward method to detect AI-generated contexts is to construct a text classification task \cite{Zellers2019DefendingAN}.
Therefore, DetectGPT \cite{Mitchell2023DetectGPTZM} introduces a strong RoBERTa-trained baseline that uses the dataset released by OpenAI \footnote{https://github.com/openai/gpt-2-output-dataset} to train a classifier.

\textbf{Model-wise Detection}
Unlike semantic-wise detection which discriminates the semantic difference between human-written and AI-generated texts, a more direct way is to explore the model-wise features.
That is, AI-generated texts show discrepancies compared with humans when fed into AI models.
These discrepancies are not easily noticed by humans as they might be a subtle difference that is possessed by a certain origin, therefore, many works focus on utilizing model-wise features such as log-likelyhood of model outputs, neural features, or bag-of-word features \cite{Bakhtin2019RealOF,Solaiman2019ReleaseSA,gehrmann-etal-2019-gltr,Jawahar2020AutomaticDO,Mitchell2023DetectGPTZM} to detect AI-generated contexts.

In the era of LLMs, there are similar works that work on using model-wise features of LLMs such as watermarking \cite{Kirchenbauer2023AWF}, and backdoor plantings \cite{Kurita2020WeightPA,Li2021BackdoorAO}.
In the computer vision field, model-wise features are also used in fake image detection \cite{Dolhansky2020TheDD} but are less related to LLM origin tracing due to the continuous nature of images.

\section{Methods}

We aim to detect whether a context is generated by an LLM system and trace the origin of the texts.
Therefore, we design a simple method \textbf{Sniffer} \footnote{This name is inspired by the fact the sniffer dogs are able to trace the scents that cannot be easily noticed by humans.} that is applicable in both white-box and black-box settings and only requires limited supervised data.

The core idea is to utilize the contrastive features between different accessible language models such as GPT-2, GPT-Neo, GPT-J, and LLaMA.
We first obtain the perplexity of a target sample $S$ based on different models denoted as $\theta_0, \theta_1,..., \theta_N$, then we craft several heuristic features and construct a simple linear classifier to classify the origin of the given sample.
Through such a feature engineering process, we can trace the origin of the target sample down to a known model $\theta_n$, we can also generalize the contrastive features to show differences between unknown source models or human-written texts.

Compared with previous detection methods, as seen in Table \ref{tab: comparison}, our proposed method is the first to allow origin tracing. Model-wise detection methods such as DetectGPT are designed to detect whether a text is generated by a certain model, which can not be well generalized to origin tracing.
Supervised learning methods such as RoBERTa finetuning, on the other hand, require a large amount of training data and are fixed to pre-defined labels, which is also limited when LLMs are developing drastically.

\begin{table}
  \caption{Different Detection Tools comparisons. Generalization ability represents how well white-box methods can detect texts generated from unknown models, therefore the supervised learning methods are not applicable in the generalization ability comparison.}
  \label{tab: comparison}
  \centering
  \begin{tabular}{llcccc}
    \toprule
    \cmidrule(r){1-5}
    Method     & Model Access & Data Request   &  Generalization & Origin Tracing \\
    \midrule
    log $p(x)$ & White-Box & Zero-Shot & Low & \xmark \\
    DetectGPT     & White-Box & Zero-Shot & Low & \xmark      \\
    Supervised-Learning     & Black-Box  & Full Data & N/A & \xmark  \\
    Sniffer & Black \& White Box & Low-Resource & High & \cmark \\ 
    \bottomrule
  \end{tabular}
\end{table}

\begin{figure}[htbp]
  \centering
{\includegraphics[width=0.9\textwidth]{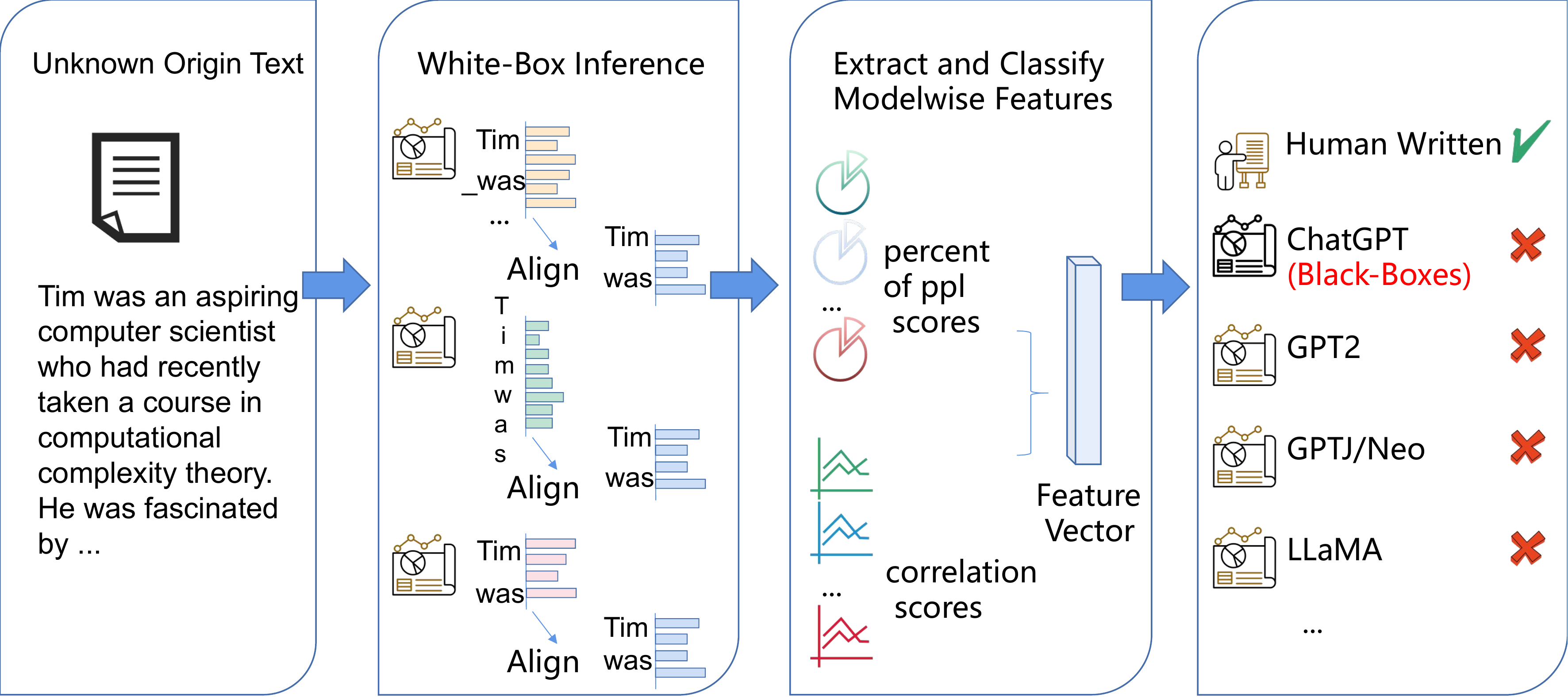}}
\vspace{-0.3cm}
  \caption{Sniffer process Illustration}
  \label{fig:sniffer}
  \vspace{-0.5cm}
\end{figure}

\subsection{Steps of Sniffer}

The process of Sniffer includes:
(1) Obtain and align token-level perplexity between different models; 
(2) Extract contrastive features;
(3) Train features for origin tracing.

\textbf{Obtain and Align Single Model Perplexity} 

Given target text $S$ and a known model $\theta_n$, we obtain the encoded tokens $\boldsymbol{x} = [x_0, x_1, ..., x_i, ...]$, the perplexity of $i^{th}$ token $x_i$ in text $S$ given model $\theta_n$ is the log-likelyhood $ll_{\theta_n}(x_i) = \text{log} {p}_{\theta_n}(x_i|x_{<i})$.
Given a list of known models $\theta_0, \theta_1,...,\theta_N$, we obtain a list of perplexities of the same text $S$.
Since the tokenization process of each LLM might be different, we use a general word-level tokenization of $S$: $\boldsymbol{w} = [w_0, w_1, ...]$ and align calculated perplexities of tokens in $\boldsymbol{x}$ to the general words.
If the word $w$ is aligned to multiple tokens in $\boldsymbol{x}$, we use the averaged perplexity; if a token in $\boldsymbol{x}$ is aligned to multiple words, we assign these words with the same value. \footnote{The details of the aligning process can be seen in the Appendix.}.
Further, the aligned perplexity is conditioned on the specific model-training process, therefore, the perplexity should be normalized for comparison between models.
We apply normalization strategies including dataset-wise-normalization and L1-normalization for these aligned features.
Dataset-wise normalization is to normalize the perplexity with an averaged perplexity on all data.
Therefore, as seen in Figure \ref{fig:sniffer}, we obtain lists of word-level perplexities $ll_{\theta_{n}}(\boldsymbol{w})$ that are aligned across models.

\textbf{Extract Contrastive Features}

After obtaining aligned perplexities $ll_{\theta_{n}}(\boldsymbol{w})$, which is a list of token-wise perplexities, we aim to search features within the list of token-wise perplexities.
Instead of introducing assumptions of the difference between human-written texts and machine-generated texts \cite{Mitchell2023DetectGPTZM}, the core idea of Sniffer is to find different features between models.
Therefore, we pose a simple hypothesis: 

\textbf{Hypothesis 1} \textit{A human written text tends to have a similar perplexity list across models, and a generated text tends to show discrepancy across models.}

To further generalize the extracted features for tracing unknown models, we pose another hypothesis:

\textbf{Hypothesis 2} \textit{The discrepancy of perplexity curves across models can reveal features of both known and unknown models.}

With these hypotheses, we can utilize the contrastive features between known models $\theta_0, \theta_1, ...$ to trace the origin of various models denoted as $\phi_0, \phi_1,...$ that includes known and unknown models. 

Specifically, given a pair-wise known model comparison between model$\theta_i$ and $\theta_j$, we have $ll_{\theta_{i}}(\boldsymbol{w})$ and $ll_{\theta_{j}}(\boldsymbol{w})$ with aligned perplexities of $L$ words in total.
We first design a \textit{percent-of-low-perplexity} score that calculates the percentage of the lower word perplexities of the perplexity of $\theta_i$ compared with $\theta_j$: 
$\text{pct}_{ij} = { \left(\sum^N_k \mathbbm{1} \left(ll_{\theta_i}(w_k) < ll_{\theta_j}(w_k) \right) \right) } / {L}$.
Here, $\mathbbm{1}(\cdot)$ is the indicator function that assigns 1 when the term is satisfied.

Given $N$ known models, we have $C_N^2$ pairs and we can obtain $C_N^2$ $\text{pct}_{ij}$ scores.
We collect all these scores as heuristic features that can be used to trace the text origins.
Plus, we also include the sentence-level perplexities and the Pearson and Spearman correlation coefficient between $ll_{\theta_{i}}(\boldsymbol{w})$ and $ll_{\theta_{j}}(\boldsymbol{w})$ to construct a feature vector as the final representation for origin tracing. For instance, given 4 known LLMs ($N=4$), we have 4 sentence-level perplexity features, $C_4^2=6$ pct-scores, and $6*2=12$ correlation coefficient score, therefore the final representation vector is a 22-dim vector.

\textbf{Train Features for Origin Tracing}

After extracting the contrastive features from different known models, we can train a simple linear classifier $F(\cdot)$ to project the extracted features to different model origins.
 To train the classifier $F(\cdot)$, we collect a small amount of human-written texts that cover various aspects and use different known models to generate texts as similar or \textit{parallel} data compared to human-written texts to train the linear classifier.

The trained classifier has several unique features:

    (1) Low-resource required: 
    Unlike semantic-wise classification tasks that require a large amount of data and a strong natural language encoder, after feature extraction, the final representation is a low-dimension vector, which requires only a small amount of data since the feature contains abundant model-wise heuristic knowledge for origin tracing;

    (2) Generalization ability to unknown models and stronger LLMs: 
    In the classifier studying process, we can collect unknown-model-generated texts to obtain their features based on known models and these features can reveal different traces compared with known models and human-written texts.
    Different from supervised-learning methods that rely on semantic-level features to detect text origins, model-wise features are \textbf{NOT} influenced as the quality of generated texts improves.
    That is, in the supervised learning methods,  stronger LLMs are harder to detect since the generated texts are human-like, but model-wise features cannot be easily optimized.
     
    (3) Extend Ability:
    In our proposed method, the linear classifier can be easily modified to trace various model origins.
    Given a new open-source model, we can easily use the feature extraction strategy above to re-train the classification model;
    given a new unknown model, we can easily collect a few generated texts and extract the features based on known models, and train a new classifier with the unknown model origin features to trace the new unknown origin.
    With such extended ability, our proposed method can be used in both black-box and white-box settings.
    

\section{Experiments}

\subsection{Dataset Construction}

In the origin tracing of LLMs, one major challenge is that LLMs are almost omniscient to world knowledge since they are trained with various and huge amounts of data.
We collect a wide range of texts from different origins for the proposed origin tracing tool Sniffer to serve as a general detector. 

We collect texts from domains including News articles, social media posts, web texts, scientific articles or academic papers, and technical documentation.
We use public datasets including XSum dataset \cite{xsum-emnlp} that contains news articles; IMDB dataset that contains social media reviews of movies; web texts \cite{radford2019language} that contains texts from common-crawled online pages; PubMed and Arxiv dataset \cite{cohan-etal-2018-discourse} that contains academic topics; Wikipedia corpus used in SQuAD dataset \cite{rajpurkar2016squad} that contains general world knowledge.
For each dataset, we randomly collect 1,000 documents and shuffle them into a human-written text dataset containing 6k documents. 
Then we generate AI-generated contexts from the texts we collected as \textit{parallel} data to study origin tracing.

When generating texts from language models such as GPT-2, we use the first 10 words as the prompt to generate a document.
When generating texts from instruction-tuned models such as GPT-3.5(text-davinci-003) and ChatGPT (turbo), we give several instructions including re-write instruction and story generation instruction to obtain a similar AI-generated text. (We show instruction details in the Appendix.)
With these simple instructions, we collect AI-generated texts from instruction-tuned LLMs.

For known origin models, we collect AI-generated texts from GPT2 (powered by OpenAI \footnote{\url{https://openai.com/}}), GPT-J and GPT-Neo (powered by EleutherAI \footnote{\url{https://www.eleuther.ai/}}), LLaMA (powered by Meta AI \footnote{\url{https://ai.facebook.com/blog/large-language-model-llama-meta-ai/}}).
For unknown origin models, we collect AI-generated texts from GPT3.5-text-davinci-003, which is an instruction-tuned model with 175B parameters.
Therefore, we collect 6k human written texts, and every 6k texts from GPT2, GPT-Neo, GPT-J, and 12k texts from GPT3 models, which is 36k texts in total.
We divide the 36k texts into a train/test split with a 90\%/10\% partition.

We name the collected dataset \textbf{SnifferBench}, which can be further used in origin tracing and AI-generated contexts detection tasks.
Further, the dataset collection process can be extended to different LLMs in the future with more different scenarios and prompts/instructions which further challenges the origin tracing ability.

\begin{figure}[htbp]
  \centering
{\includegraphics[width=1\textwidth]{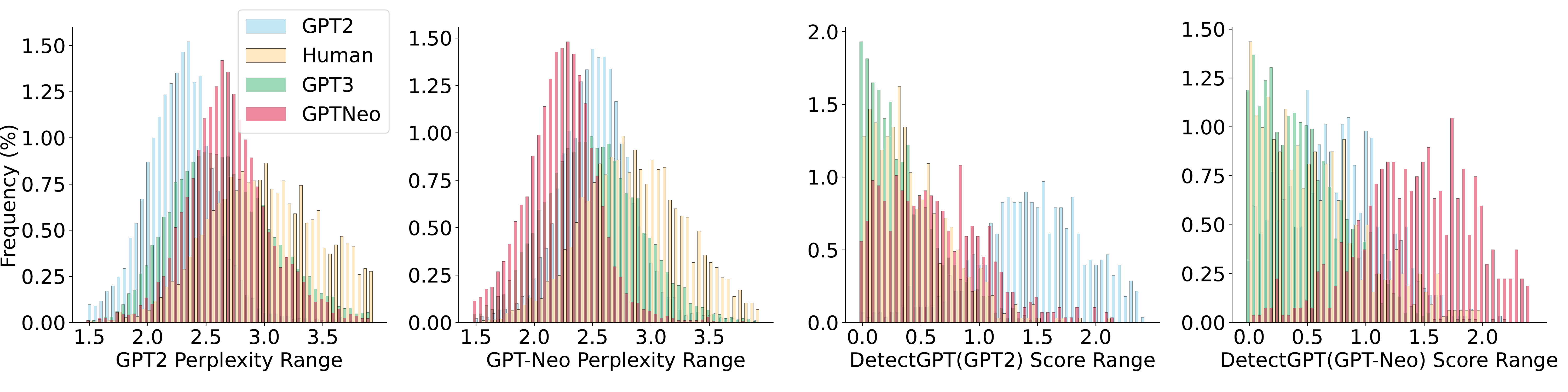}}
\vspace{-0.5cm}
  \caption{The discrepancy between different text origins in different baseline methods. In each figure, different bars show different text origins and each figure is to use a certain model of a certain detect method to test given texts.}
  \label{fig:figs}
  \vspace{-0.5cm}
\end{figure}

\subsection{Baseline Methods}

As we compare Sniffer with previous detection methods in the form of the methods, we construct experiments to explore how previous methods fail to run origin tracing. 

We first implement a $\text{log}p(x)$ method which is also used in the original version of GPTZero \footnote{\url{https://gptzero.me/}};
then we implement the DetectGPT method which introduces a discrepancy score to discriminate AI-generated texts by adding multiple perturbations (we try 40 perturbations for each sample).
We collect the sentence-level perplexity and the DetectGPT discrepancy score of the collected dataset based on a certain known model $\theta_n$ and draw a histogram showing the score distributions of different origin texts.
Then we select a threshold manually as the discrimination boundary in $\text{log}p(x)$ and DetectGPT.
In Figure \ref{fig:figs}, we plot the histogram of the discrepancy between different texts' origins in $\text{log}p(x)$ method and DetectGPT method.
In $\text{log}p(x)$, we plot the perplexity score of different text origins using a specific model such as GPT-2 or GPT-Neo, and in the DetectGPT method, we use their proposed z-score.
As seen in Figure \ref{fig:figs}, though there are multiple peaks showing that there are differences between different text origins, the overlap is too large to successfully separate different text origins.
we can conclude that it is extremely difficult to discriminate text origins from $\text{log}p(x)$ features or z-score features used in DetectGPT.
Therefore, it is important to introduce strong features to trace the texts' origins.

\subsection{Implementations of Sniffer}

In Sniffer, we select several open-source (L)LMs as known models:
we use GPT2-xl(1.5B), GPT-Neo(2.7B), GPT-J(6B) and LLaMA(7B) as known models.
In the SnifferBench, we collect texts from origins including GPT2(OpenAI), GPT-Neo and GPT-J(EleutherAI), LLaMA (MetaAI), ChatGPT(GPT3.5-turbo from OpenAI), and human-written texts, 
therefore, the unknown origin is the ChatGPT(GPT3.5-turbo) model since they are not open-source models.
The goal of origin tracing in Sniffer is to trace both known and unknown origins.
We construct an inference server for each known model based on NVIDIA4090 GPUs and set the max sequence length given the maximum GPU allowance. 
We align all texts with a white-space tokenizer to obtain uniform tokenizations.

Besides Sniffer which uses 4 known models and utilizes a linear classifier for origin tracing, we introduce a new variant of Sniffer: \textbf{Sniffer(+GPT3)}
As we are able to obtain logits of the generated texts in the GPT-3 API provided by OpenAI, we are able to treat GPT-3 model as a white-box model.
Therefore, we collect a subset in the SnifferBench to test the origin tracing with GPT-3 models as white boxes.
Therefore, in our implementation of Sniffer, we have 5 models in total which result in a 35-dimension vector in obtaining the sniffer feature.
Here, we align the tokens with the tokenizer provided by OpenAI.

\subsection{Metrics of Origin Tracing}

In the SnifferBench, we collect texts from different origins,
therefore, we calculate the precision and recall of each text origin tracing prediction.
We can only calculate the known model and the human-written texts' precision and recall in baseline methods since these methods cannot be used in the origin tracing task.

\subsection{Results of Sniffer Origin Tracing}

In Table \ref{tab:main-results}, we can observe that model-wise detection methods such as perplexity-based method $\text{log}p(x)$ and perturbation-based method DetectGPT cannot be used in origin tracing task.
When predicting whether the texts are generated by the specific model GPT-2, $\text{log}p(x)$ and DetectGPT can obtain reasonable results but fail to discriminate the mixture of texts generated by EleugherAI-released models including GPT-J and GPTNeo, not to mention that these methods cannot discriminate other models.

We can observe that Sniffer can obtain impressive results in tracing the text origins in the known models including GPT-2, GPT-J/Neo;
further, when Sniffer does not use the GPT-3 models as white boxes, Sniffer can generalize to trace GPT-3 model origins and can properly discriminate them from human-written lines.
Such an ability cannot be easily obtained as the GPT-3 models are stronger and similar to humans, meanwhile remaining unknown to the public.

Plus, we can observe that recently released LLaMA models are more difficult to trace, though it is a white box in Sniffer, indicating that stronger LLMs are harder to detect.
When LLMs are being studied by a wide range of researchers, it is also important to be alert that stronger models may be harder to detect, and harder to control which can cause potential harm to society.

\begin{table}
  \caption{Accuracy/Recall Results of Origin Tracing. Since $\text{log}p(x)$ and DetectGPT methods are unable to trace different text origins, only the corresponding model-generated text and human-written text detection results are listed. $\%$ in Sniffer is to train Sniffer with limited training data; $\text{log}p(x)$ and pct-score -only are the variants that only use the corresponding features for the linear projection.}
  \label{tab:main-results}
  \centering
  \small
  \begin{tabular}{lcccccccccc}
    \toprule

    \multirow{2}{*}{Method}     & \multicolumn{4}{c}{Different Text Origins}  \\
    \cmidrule(r){2-7}
    & GPT-2  & GPT-J/Neo  & LLama  & ChatGPT & Human & Total \\
    & (OpenAI) & (EleutherAI) & (Meta) & (OpenAI) & \\
    \midrule

    log $p(x)$ (GPT-2) & 80.9/89.3  & - & - &- & 87.9/78.7 & - \\
    log $p(x)$ (GPT-J)  & - & 71.7/78.9  &- & -  & 76.3/68.5 & - \\
    log $p(x)$ (GPT-Neo) & - & 78.4/84.9 & - &- & 83.3/76.4 & - \\
    DetectGPT (GPT-2) & 88.9/88.9  & - & - &- & 89.9/90.2 & - \\
    DetectGPT (GPT-J) & - & 74.4/79.3 & - &- & 80.0/75.5 & - \\
    DetectGPT (GPT-Neo) & - & 81.2/87.5 & - &- & 87.8/81.9 & - \\
    \midrule
    Sniffer & 98.7/{96.9} & {96.6}/\textbf{98.0} & \textbf{85.0}/\textbf{84.3} & \textbf{77.7}/{82.3} & {68.1}/{60.3} & \textbf{86.0}/- \\
    Sniffer ($10\%$)  & 97.3/96.3 & \textbf{96.7}/96.1 & 80.9/77.2 & 73.9/77.3 & 58.9/\textbf{67.7} & 82.6/- \\
    Sniffer ($5\%$)   & 97.3/97.5 & 96.6/95.1 & 76.1/74.0 & 71.4/76.7 & 58.8/53.4 & 81.3/- \\
    Sniffer ($1\%$)   & 97.9/94.4 & 91.0/95.2 & 65.8/60.2 & 67.4/76.3 & 60.0/46.4 & 77.7/- \\
    Sniffer (L1-norm) & 97.8/\textbf{98.3} & \textbf{96.7}/95.9 & 75.2/74.4 & 74.7/82.4 & \textbf{75.7}/62.4 & 84.1/- \\
    Sniffer ($\text{log}p(x)$ only) & \textbf{98.9}/97.7 & 94.1/94.8 & 60.4/49.3 & 64.6/78.8 & 63.0/47.6 & 77.3/- \\
    Sniffer (pct-score only)        & 98.3/96.6 & 94.0/94.8 & 59.5/53.2 & 60.0/79.9 & 58.3/26.8 & 75.1/- \\
    Sniffer ($\text{log}p(x)$ + pct-score) & 98.6/97.2 & 96.5/96.2 & 69.6/65.0 & 71.0/\textbf{82.5} & 66.3/51.1 & 81.4/- \\
    
    \bottomrule
  \end{tabular}
\end{table}

\begin{table}
  \caption{Results of Sniffer(+GPT3): We use GPT-3.5(text-davinci-003) to replace ChatGPT(turbo) since ChatGPT(turbo) does not provide logits.}
  \label{tab:white-box-gpt}
  \centering
  \small
  \begin{tabular}{lcccccccccc}
    \toprule

    \multirow{2}{*}{Method}     & \multicolumn{4}{c}{Different Text Origins}  \\
    \cmidrule(r){2-7}
    & GPT-2  & GPT-J/Neo  & LLama  & GPT-3 & Human & Total \\
    & (OpenAI) & (EleutherAI) & (Meta) & (OpenAI) & \\
    \midrule
    Sniffer($1\%$) & \textbf{93.8}/{90.0} & 89.6/88.0 & 59.1/47.4 & 57.6/59.6 & 35.3/42.9 & 68.1/- \\
    Sniffer(+GPT3)($1\%$) & {92.0}/\textbf{91.8} & \textbf{90.0}/\textbf{92.4} & \textbf{81.6}/\textbf{72.6} & \textbf{70.0}/\textbf{65.6} & \textbf{42.1}/\textbf{50.0} & \textbf{76.0}/- \\
    
    \bottomrule
  \end{tabular}
\end{table}

\subsection{Different Sniffers}

\textbf{Sniffer+GPT3}

In Table \ref{tab:white-box-gpt}, we list the results of using Sniffer(+GPT3) to trace the origins of texts generated by GPT3 models while we treat GPT3 models as white boxes.
We use the OpenAI service that returns token logits therefore we use the text-davinci-003 model and test on a $1\%$ dataset compared with the full dataset.
Here, we replace texts generated from ChatGPT with davinci-003 outputs therefore the testset is different from the one used in tracing ChatGPT texts.

As seen, when the Sniffer feature can use GPT-3 logits, the results grow significantly higher compared with generalizing features from GPT2/J/Neo and LLaMA models to trace GPT-3.
Therefore, we can conclude that although Sniffer is able to generalize its features to trace unknown origins, it is better to possess the LLMs, indicating that the actual LLM providers must be more cautious when releasing LLMs since they are more capable to avoid abuse or malicious usage of LLMs.

\textbf{Few-shot Experiments}

As we illustrated, in methods using model-wise features, we can use limited data to construct a powerful detector.
Therefore, we use different numbers of training data to train Sniffer.
As seen, when we use limited data to train the Sniffer model, the performances are not significantly harmed, indicating that the model-wise features are high-quality features that reveal obvious traces of texts for the origin tracing classification.

\textbf{Feature Ablations}

In the extracted features, we observe that using L1-norm can help obtain a higher performance in tracing human texts, but is rather weak in tracing different model origins, especially when tracing LLaMA. Therefore, we use datase-wise norm for the rest experiments.

To further analyze how the extracted features help trace the text origins, we run a simple ablation test that uses different features proposed in Sniffer.
As seen in Table \ref{tab:main-results}, the perplexity score $\text{log}p(x)$ is one key metric but can be significantly improved by the percent-of-perplexity score (pct-score), showing that we can trace text origins by analyzing the discrepancies between models when testing same texts as Sniffer did.

\begin{figure}[]
  \centering
{\includegraphics[width=0.9\textwidth]{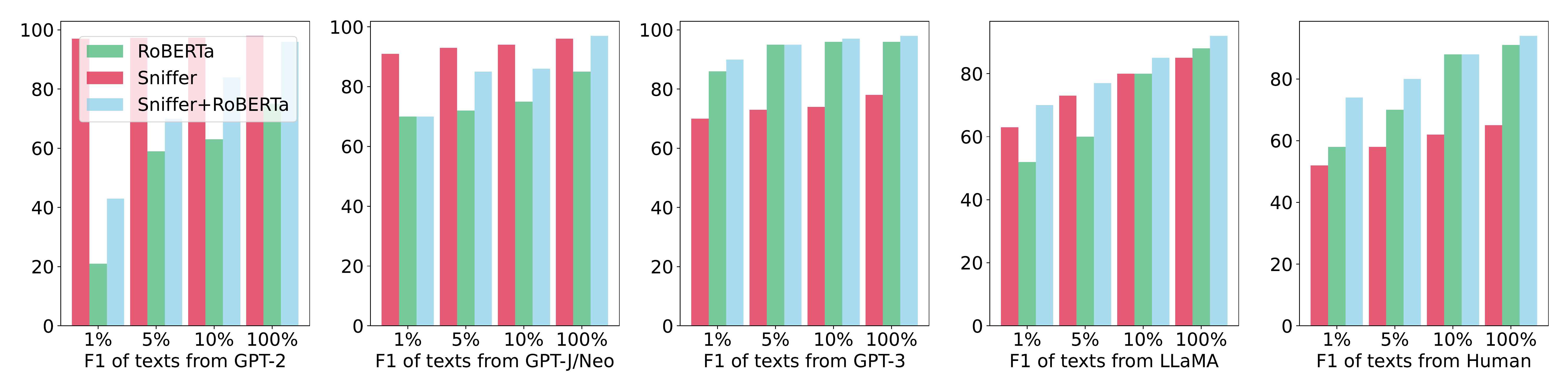}}
  \vspace{-0.5cm}
  \caption{Comparison with Supervised Learning Methods that utilize semantic-wise features.}
  \label{fig:supervised-learning}
  \vspace{-0.5cm}
\end{figure}

\textbf{Compare with Supervised Learning Methods}

As discussed, the model-wise features trace the text origins by analyzing the model discrepancies, and the semantic-wise features are used as a fine-tuning task.
We compare Sniffer with a fine-tuned RoBERTa model and measure the f1-score.
Further, we combine Sniffer features and RoBERTa \texttt{[CLS]} feature to train a linear classifier as Sniffer+RoBERTa to test the performances of combinations of model-wise and semantic-wise features.
As seen in Figure \ref{fig:supervised-learning}, the supervised learning method fails to solve the problem when the data is limited.
Still, when the data is abundant, the performances are stronger than model-wise features,
which is also discussed in \citet{Mitchell2023DetectGPTZM,2023OnTP}. 

We need to notice that the supervised learning method fails to detect GPT-2/GPT-J/Neo generated texts, which is different from Sniffer results as GPT-2 texts are known to be less fluent compared with stronger models such as GPT-3 models.
Therefore, we can assume that the RoBERTa models may capture some specific patterns since the GPT-3 generated texts only use several static instructions and the semantic features are easier to detect.
The failure of GPT-2 detection indicates that semantic features are limited in origin tracing, calling for better algorithms to utilize model-wise features.
In Sniffer+RoBERTa, we obtain promising results in tracing all origins, indicating that a proper combination of two types of features can help better trace text origins.

\begin{figure}[]
  \centering
{\includegraphics[width=1\textwidth]{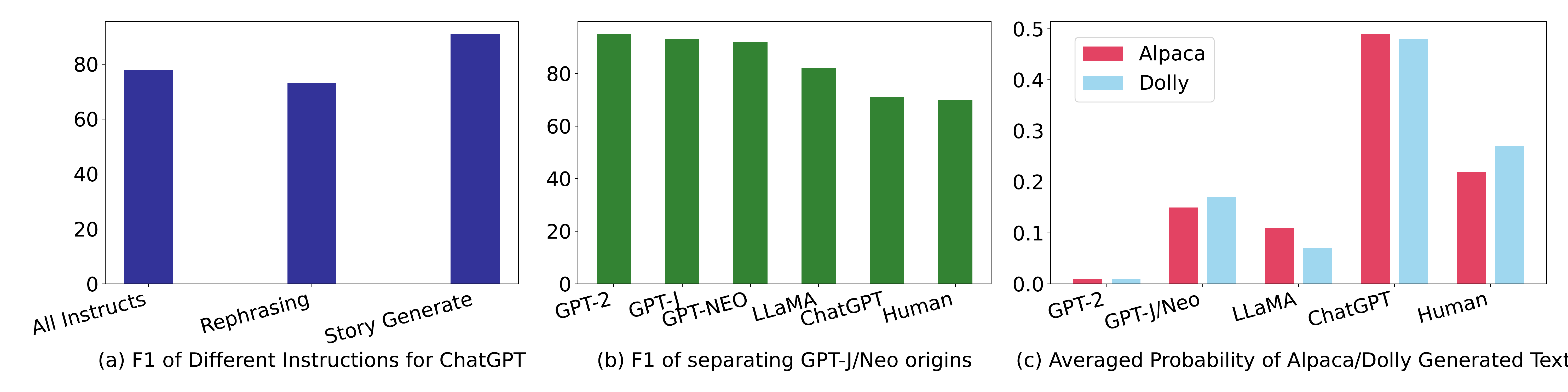}}
\vspace{-0.5cm}
  \caption{Tracing different types of generated texts: (a) plots the ChatGPT tracing results that use different instructions; (b) plots the tracing results that separate GPT-J and GPT-Neo origins and calculate the f1-score of corresponding origins;
  (c) plots the tracing results that test Alpaca and Dolly models that use ChatGPT instructions to supervise fine-tuning the LLaMA/GPT-J models to build an instructed LLM.}
  \label{fig:mix}
  \vspace{-0.5cm}
\end{figure}

\subsection{Difficulty of Tracing Different Types of Generated Texts}

\textbf{Generation Genre Influence}

As illustrated, we use several different instructions to instruct GPT3.5 (turbo) models and GPT3.5-text-davinci-003 models, we further discuss the tracing difficulty when the texts are generated by different instructions.

As seen in Figure \ref{fig:mix}(a), we find that when the texts are generated by rephrasing instructions, the texts are rather easy to trace while texts generated from a summary are harder to trace.
Such an observation indicates that the difficulty of tracing texts generated by strong LLMs is also different when the instructions are different.
The rephrased texts are more similar to human-written texts, therefore, are more difficult to detect.

\textbf{Mixing Origin Tracing}

In our origin tracing experiment setup, we divide GPT-J/Neo generated texts into the same origin since these models are provided by the same facility.
That is, LLM origins can have different levels of classifications.
As LLMs can be trained with \textit{distill texts} from strong LLMs such as ChatGPT, it is harder to trace the origin if the model is trained based on a base model such as GPT-J or LLaMA but is instructed by ChatGPT-generated outputs.
It is hard to tell whether the base model or the instruct model has a larger impact on the mixed-origin model.
Therefore, we construct experiments to first separate GPT-J/Neo origins using the proposed datasets.
Further, we test mixed-origin models such as Alpaca \cite{alpaca} (ChatGPT instructions tuned based on LLaMA) and Dolly \footnote{\url{https://huggingface.co/databricks/dolly-v2-12b}} (ChatGPT instructions tuned based on GPT-J) by instructing these models to generate 400 samples and testing them with Sniffer.


In Figure \ref{fig:mix}(b), we show the f1-score of origin tracing that separates GPT-J and GPT-Neo models using the full data of SnifferBench.
As seen, Sniffer is able to successfully divide GPT-J and GPT-Neo, indicating that the text origin divide can be of different levels.
We can trace the text origins of a specific model or a certain party.

In Figure \ref{fig:mix}(c), we list the averaged probability of the Sniffer inference results of supervise fine-tuned model Alpaca and Dolly.
As seen, texts from both Alpaca and Dolly tend to be categorized as ChatGPT-generated texts, indicating that the align process has a more significant impact on the generated texts compared with the base model. 
Therefore, Sniffer can be used as a detector for testing whether a model is trained from ChatGPT-generated instructions, helping protect the originality of LLMs.



\section{Conclusion and Future Work}

In this paper, we first introduce the concept of origin tracing, an important direction in the era of LLMs.
Then we discuss two lines of AI-generated context detection and origin tracing methods and point out the necessity of studying model-wise features for origin tracing.
We further design a simple Sniffer method as well as a benchmark to test the origin tracing challenge.
Through extensive experiments, we find that the current origin tracing field is full of challenges including tracing texts from models with mixed origins; combining semantic features and model-wise features; tracing texts from LLMs that are given various instructions.
Therefore, we can hope for a continuous line of works that study the origin tracing of LLMs and hope to improve the trustworthy and safe usage of LLMs.

{
\small
\bibliography{neurips_2022}
}

\clearpage

\appendix

\section{Appendix}


\begin{figure}[]
  \centering
{\includegraphics[width=1\textwidth]{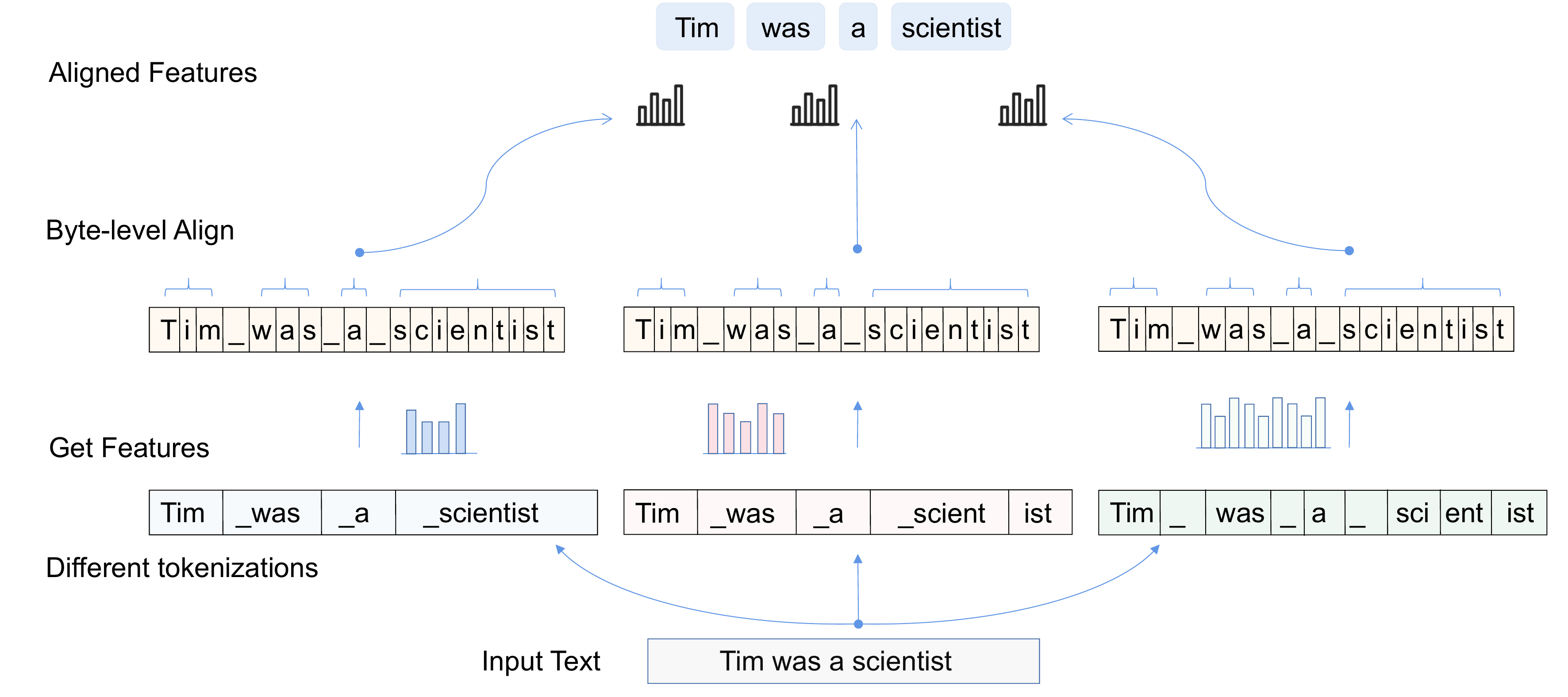}}
  \caption{Align Strategies}
  \label{fig:align}
\end{figure}

\textbf{Align Strategies}

As mentioned, we align tokens with different tokenization methods with a uniform tokenization strategy to compare the perplexity at the same level.
As shown in Figure \ref{fig:align}, when the texts are tokenized by different tokenizers, we project different tokens to the byte level, and from the byte-level projections, we align different tokenized texts to a list of tokens with the same tokenizations.

\begin{table}
  \caption{Accuracy/Recall results of Chinese Dataset Performances of Origin Tracing.}
  \label{tab:cn-results}
  \centering
  \scriptsize
  \begin{tabular}{lcccccccccc}
    \toprule

    \multirow{2}{*}{Method}     & \multicolumn{4}{c}{Different Text Origins}  \\
    \cmidrule(r){2-7}
    & Wenzhong  & Damo  & Skytext  & ChatGLM & ChatGPT & MOSS & Human & Total \\
    & Fengshenbang & Damo & SkyWork & THUDM & OpenAI & FudanNLP & \\

    \midrule
    Sniffer  & 85.5/89.0 & 89.4/89.8 & 95.2/95.5 & 81.5/85.7 & 75.9/82.0  & 69.7/50.4 &  60.1/56.4 & 79.8/- \\
    
    \bottomrule
  \end{tabular}
\end{table}

\textbf{Chinese Corpus Experiments}

We also conduct a Chinese version of Sniffer as well as a collected testset for studying different types of LLMs.

We sample 1k samples from each open-source dataset of various domains including a long-document corpus of stories \cite{guan2022lot}; various web texts from Iflytek classification dataset \cite{xu-etal-2020-clue}; Chinese academical documents from CSL dataset \cite{li-etal-2022-csl}; Chinese Wikipedia corpus from CMRC dataset \cite{cui-etal-2019-span}; review corpus from Xiecheng APP \footnote{\url{https://huggingface.co/datasets/seamew/ChnSentiCorp}}.
We collect these datasets that cover various domains with relatively long documents (more than 200 Chinese characters per document).
The dataset construction setup is similar to English dataset setups.
We also use ChatGPT (GPT turbo) model to generate Chinese texts with the Chinese instructions as ChatGPT (GPT turbo) is a strong multi-lingual model.

As for open-source model selections, we adopt several open-source Chinese LLMs including Wenzhong model \cite{fengshenbang}, a 2.7B GPT-2 style Chinese LLM; Damo model \footnote{\url{https://modelscope.cn/models/damo/nlp_gpt3_text-generation_2.7B/summary}}, another 2.7B GPT-2 style Chinese LLM; Skytext model \footnote{\url{https://huggingface.co/SkyWork/SkyTextTiny}}, a 3B chatbot, and ChatGLM \cite{zeng2023glm-130b}, a 6B ChatGPT-style model trained with instructions. 

For the generalization test of black-box models, we use ChatGPT (GPT turbo) and MOSS, a 16B Chinese LLM \footnote{\url{https://github.com/OpenLMLab/MOSS}} with instructions that ask LLMs to re-write the given document as black-box origin tests.

We generate Chinese AI-generated contexts with open-source models and black-box models to build the whole Chinese benchmark for origin tracing tests.

As seen in Table \ref{tab:cn-results}, the Chinese texts show similar performances with English texts, indicating that the origin tracing strategy can be used in different languages.

\begin{table}[]
\caption{Details of instructions used in Generating texts from ChatGPT, GPT3.5(text-davinci-003), Alpaca and Dolly models. In the story-generation, we use two instructs to obtain the generated document.}
  \label{tab:instruct}
    \centering
    \setlength{\tabcolsep}{6pt}
    \small
    \begin{tabular}{lll}
    \toprule
    \multirow{2}{*}{Instruction Type} & \multirow{2}{*}{Models} & \multirow{2}{*}{Instructions} \\
     \\
    \midrule
    Re-write & ChatGPT & \makecell[l]{You are an assistant that can rephrase the provided content.\\ rephrase the following content: [document]}\\
    \midrule
    Story-Gen & ChatGPT & \makecell[l]{\textit{Instruct 1:} You are an assistant that can summarize the provided content. \\ summarize the following content: [document] \\ \textit{Instruct 2:} You are an assistant that can write a natural and fluent document\\ based on the provided content, do not use fixed or academic writing style\\ write a natural and fluent document like human \\based on the following content: [summary]} \\
    \midrule
    Re-write & GPT3.5 & Rephrase the following content: [document] \\
    \midrule
    Re-write & \makecell[l]{Alpaca\\ \& \\Dolly} & \makecell[l]{
        Below is an instruction that describes a task, paired with an input that \\ provides further context. Write a response that appropriately completes \\the request.\\
        \\
        \#\#\# Instruction:\\
        Rewrite the following paragraph in a different style using your own words.\\
        \\
        \#\#\# Input:\\
        \text{[document]}\\
        \\
    }\\
    
    \bottomrule
    \end{tabular}

\end{table}

\textbf{Instruction Details}

\textbf{Case Studies}

We list several case studies that are randomly selected from the testset.
We show the texts to be detected, percent-of-perplexity scores, correlation scores calculated by Sniffer, and the tracing results.

As listed below, we show the list of perplexity scores of the given text and the extracted features, then we show the predicted tracing result.
As seen, the perplexity list shows a similar trend between different models, but the perplexity value tends to show differences across models.
For instance, in the GPT-2 generated text calculated list of perplexity, the GPT-2 perplexity is relatively lower than other perplexities calculated by other models, revealing the feature that helps trace the GPT-2 texts.
On the other hand, for human-written texts, and texts generated by strong LLMs such as ChatGPT, the value does not show obvious differences between different models, for instance, some peaks in the perplexity list can be from the LLaMA models and some can be from some other models, which supports the hypothesis made above.

\begin{table}[]
    \centering
    \setlength{\tabcolsep}{6pt}
    \small
    \begin{tabular}{lll}
    \toprule
    \multirow{2}{*}{Text Origin} & \multirow{2}{*}{} \\
     \\
    \midrule
    
    \multirow{5}{*}{GPT-2} & \makecell[l]{
       I thought that Mukhsin has been wonderfully written. Its not a work of fiction, but it certainly \\ feels like a novel. I agree that there is much to admire about, despite flaws, which will be \\ evident to many of its readers. \\
    } \\
    & \includegraphics[width=0.8\textwidth]{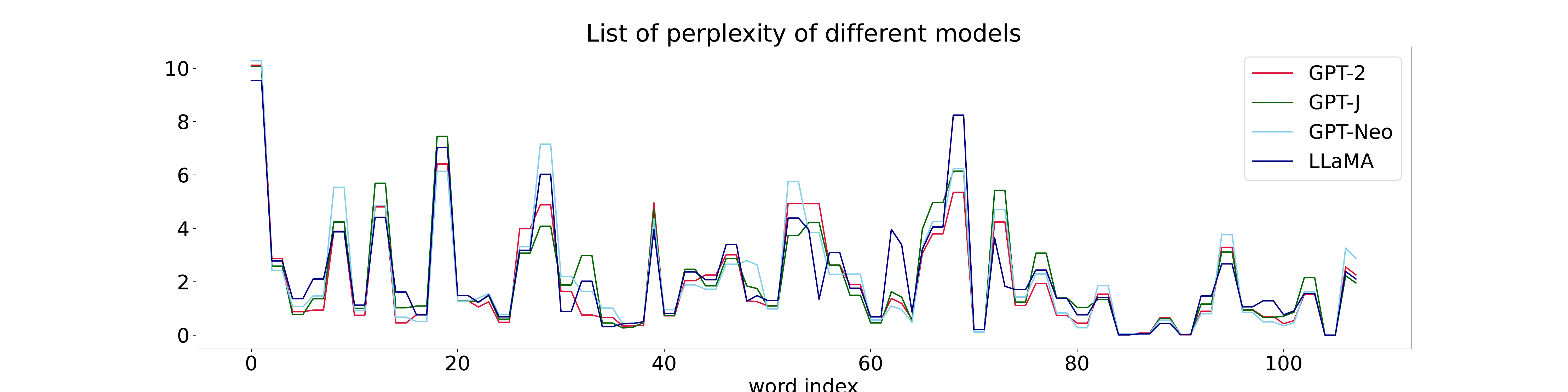} \\
    \cline{2-2}
    &  
    \multicolumn{1}{l}{
         \begin{tabular}{lll}\makecell[l]{Extracted Features: \\
           log $p(x)$ (GPT-2):  2.04 \\
           log $p(x)$ (GPT-Neo):2.18 \\
           log $p(x)$ (GPT-J):  2.18 \\
           log $p(x)$ (LLaMA):  2.21
         }
         \end{tabular}
    }\\
     & 
     \multicolumn{1}{l}{
        \begin{tabular}{lll}
           \multirow{6}{*}{pct-scores / Pearson scores / Spearman scores: }
            &  GPT-2 and GPT-Neo &  0.59/0.97/0.97   \\
            &  GPT-2 and GPT-J &    0.57/0.96/0.95   \\
            &  GPT-2 and LLaMA &    0.63/0.92/0.91   \\
            &  GPT-Neo and GPT-J &  0.56/0.94/0.94   \\
            &  GPT-Neo and LLaMA &  0.52/0.92/0.89   \\
            &  GPT-J and LLaMA &    0.55/0.92/0.93   \\
         \end{tabular}
     }
      \\
    \cline{2-2}
    & 
    \makecell[l]{Tracing Results: \\
        GPT-2: \textcolor{teal}{99.8 \%} \\
        GPT-J/Neo: 0.1 \% \\
        LLaMA: 0.0\% \\
        ChatGPT: 0.1 \% \\
        Human: 0.0 \%
     } \\
     
    \midrule
    
    \multirow{5}{*}{GPT-Neo} & \makecell[l]{
       Ten passengers on board the bus were reported to have either been injured, or have since been \\ released from hospital. "This is an extremely rare occurrence," said an unnamed source at the \\ scene. A second group of tourists were on the bus at the time of the incident but were not \\ injured, according to reports. \\
    } \\
    & \includegraphics[width=0.8\textwidth]{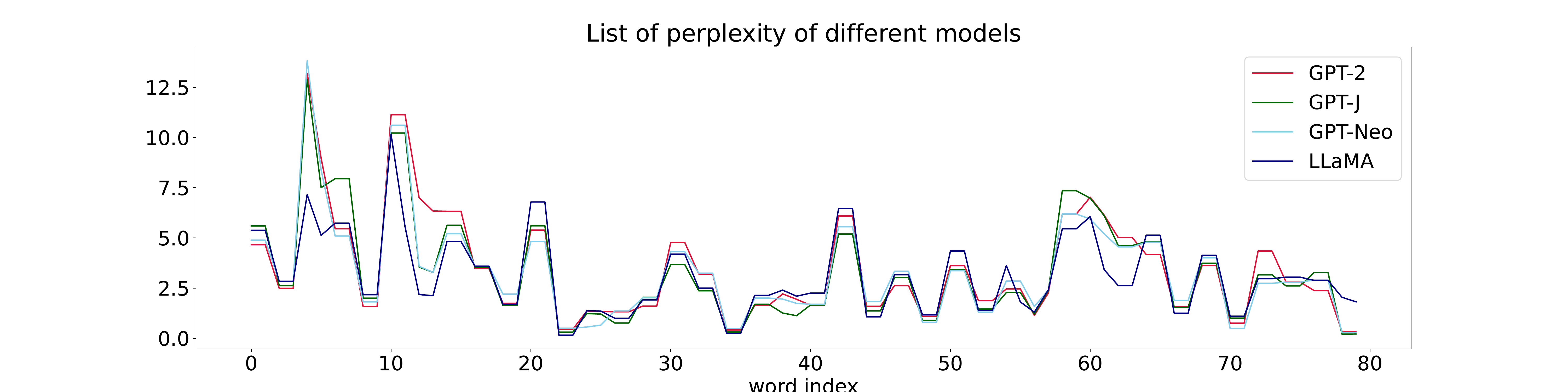} \\
    \cline{2-2}
    &  
    \multicolumn{1}{l}{
         \begin{tabular}{lll}\makecell[l]{Extracted Features: \\
           log $p(x)$ (GPT-2):  3.43 \\
           log $p(x)$ (GPT-Neo):3.21 \\
           log $p(x)$ (GPT-J):  3.23 \\
           log $p(x)$ (LLaMA):  3.08
         }
         \end{tabular}
    }\\
     & 
     \multicolumn{1}{l}{
        \begin{tabular}{lll}
           \multirow{6}{*}{pct-scores / Pearson scores / Spearman scores: }
            &  GPT-2 and GPT-Neo &  0.54/0.96/0.95   \\
            &  GPT-2 and GPT-J &    0.40/0.95/0.95   \\
            &  GPT-2 and LLaMA &    0.56/0.83/0.87   \\
            &  GPT-Neo and GPT-J &  0.42/0.97/0.98   \\
            &  GPT-Neo and LLaMA &  0.49/0.86/0.91   \\
            &  GPT-J and LLaMA &    0.56/0.89/0.93   \\
         \end{tabular}
     }
      \\
    \cline{2-2}
    & 
    \makecell[l]{Tracing Results: \\
        GPT-2: 0.3\% \\
        GPT-J/Neo: \textcolor{teal}{72.6 \%} \\
        LLaMA: 3.0\% \\
        ChatGPT: 18.3 \% \\
        Human: 5.8 \%
     } \\

    \bottomrule
    \end{tabular}
    \caption{Random selected case studies (a).
    }
    \label{tab:casestudy-a}
\end{table}

\begin{table}[]
    \centering
    \setlength{\tabcolsep}{6pt}
    \small
    \begin{tabular}{lll}
    \toprule
    \multirow{2}{*}{Text Origin} & \multirow{2}{*}{} \\
     \\
    \midrule
    \multirow{5}{*}{GPT-J} & \makecell[l]{
       He said ministers who disagreed with austerity measures could not go to other countries for \\ talks as Germany would only make life difficult for them. And he said that he was a “proud \\ German”. \\
    } \\
    & \includegraphics[width=0.8\textwidth]{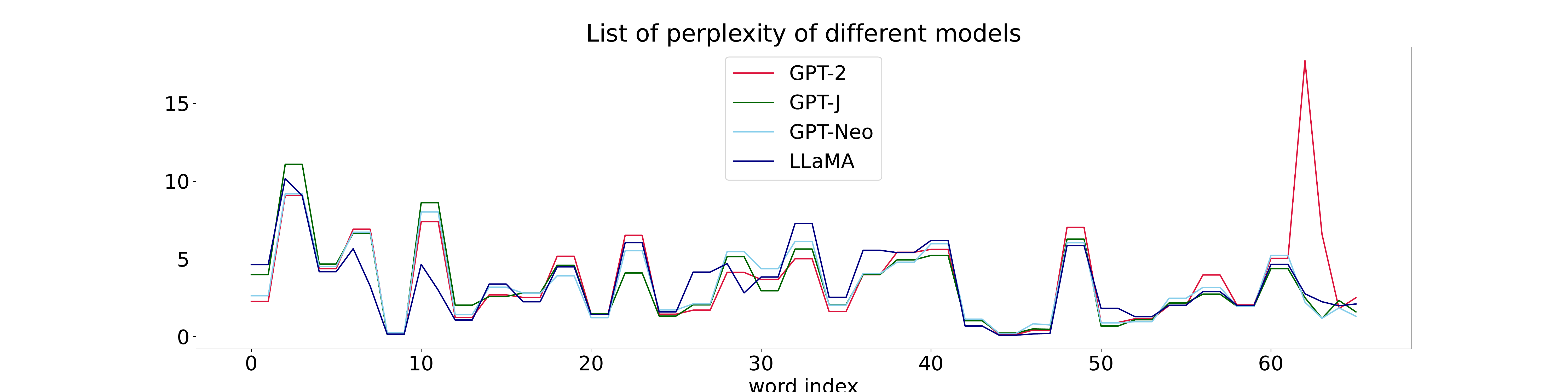} \\
    \cline{2-2}
    &  
    \multicolumn{1}{l}{
         \begin{tabular}{lll}\makecell[l]{Extracted Features: \\
           log $p(x)$ (GPT-2):  3.86 \\
           log $p(x)$ (GPT-Neo):2.99 \\
           log $p(x)$ (GPT-J):  2.96 \\
           log $p(x)$ (LLaMA):  3.07
         }
         \end{tabular}
    }\\
     & 
     \multicolumn{1}{l}{
        \begin{tabular}{lll}
           \multirow{6}{*}{pct-scores / Pearson scores / Spearman scores: }
            &  GPT-2 and GPT-Neo &  0.65/0.70/0.90   \\
            &  GPT-2 and GPT-J &    0.47/0.70/0.89   \\
            &  GPT-2 and LLaMA &    0.44/0.65/0.85   \\
            &  GPT-Neo and GPT-J &  0.44/0.96/0.98   \\
            &  GPT-Neo and LLaMA &  0.50/0.86/0.90   \\
            &  GPT-J and LLaMA &    0.56/0.85/0.89   \\
         \end{tabular}
     }
      \\
    \cline{2-2}
    & 
    \makecell[l]{Tracing Results: \\
        GPT-2: 0.0\% \\
        GPT-J/Neo: \textcolor{teal}{100.0 \%} \\
        LLaMA: 0.0\% \\
        ChatGPT: 0.0 \% \\
        Human: 0.0 \%
     } \\
    
    \midrule
    
    \multirow{5}{*}{LLaMA} & \makecell[l]{
       This is a very memorable spaghetti western. It has a great storyline about two brothers, one \\ a sheriff and the other a gunfighter. It is a very good one. I will recommend it to my friends. \\
    } \\
    & \includegraphics[width=0.8\textwidth]{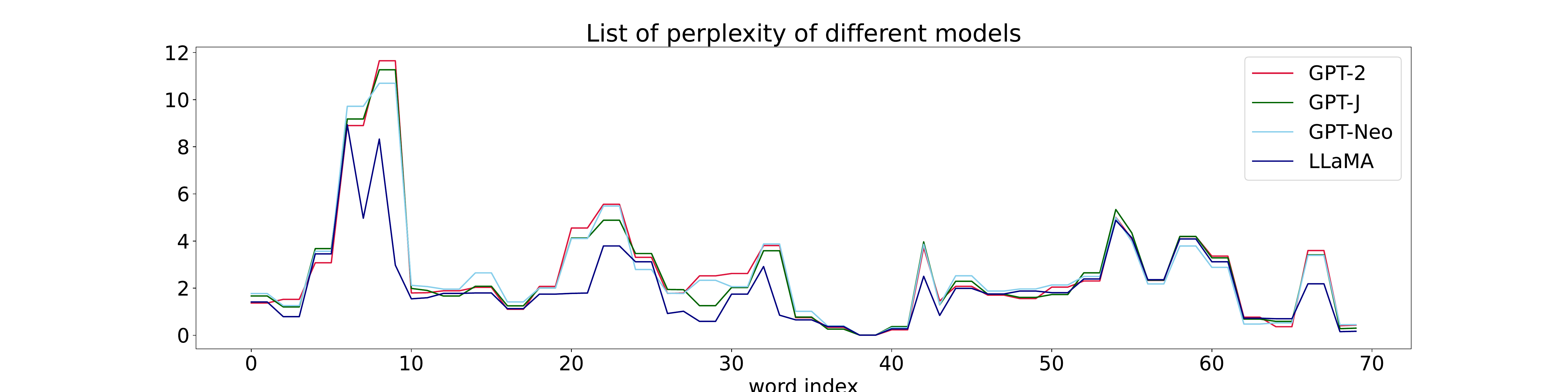} \\
    \cline{2-2}
    &  
    \multicolumn{1}{l}{
         \begin{tabular}{lll}\makecell[l]{Extracted Features: \\
           log $p(x)$ (GPT-2):  2.48 \\
           log $p(x)$ (GPT-Neo):2.46 \\
           log $p(x)$ (GPT-J):  2.42 \\
           log $p(x)$ (LLaMA):  1.91
         }
         \end{tabular}
    }\\
     & 
     \multicolumn{1}{l}{
        \begin{tabular}{lll}
           \multirow{6}{*}{pct-scores / Pearson scores / Spearman scores: }
            &  GPT-2 and GPT-Neo &  0.57/0.99/0.97   \\
            &  GPT-2 and GPT-J &    0.56/0.99/0.95   \\
            &  GPT-2 and LLaMA &    0.34/0.83/0.85   \\
            &  GPT-Neo and GPT-J &  0.47/0.99/0.96   \\
            &  GPT-Neo and LLaMA &  0.21/0.85/0.87   \\
            &  GPT-J and LLaMA &    0.23/0.86/0.91   \\
         \end{tabular}
     }
      \\
    \cline{2-2}
    & 
    \makecell[l]{Tracing Results: \\
        GPT-2: 0.2 \% \\
        GPT-J/Neo: 0.3 \% \\
        LLaMA: \textcolor{teal}{59.1 \%} \\
        ChatGPT: 31.3 \% \\
        Human: 9.0 \%
     } \\

    \bottomrule
    \end{tabular}
    \caption{Random selected case studies (b).
    }
    \label{tab:casestudy-b}
\end{table}

\begin{table}[]
    \centering
    \setlength{\tabcolsep}{6pt}
    \small
    \begin{tabular}{lll}
    \toprule
    \multirow{2}{*}{Text Origin} & \multirow{2}{*}{} \\
     \\
    \midrule
    \multirow{5}{*}{Chat-GPT} & \makecell[l]{
       The film in question focuses on the character played by the immensely talented Helena \\ Bonham Carter, who delivered an impressive performance despite being confined to a \\ wheelchair throughout the movie. \\
    } \\
    & \includegraphics[width=0.8\textwidth]{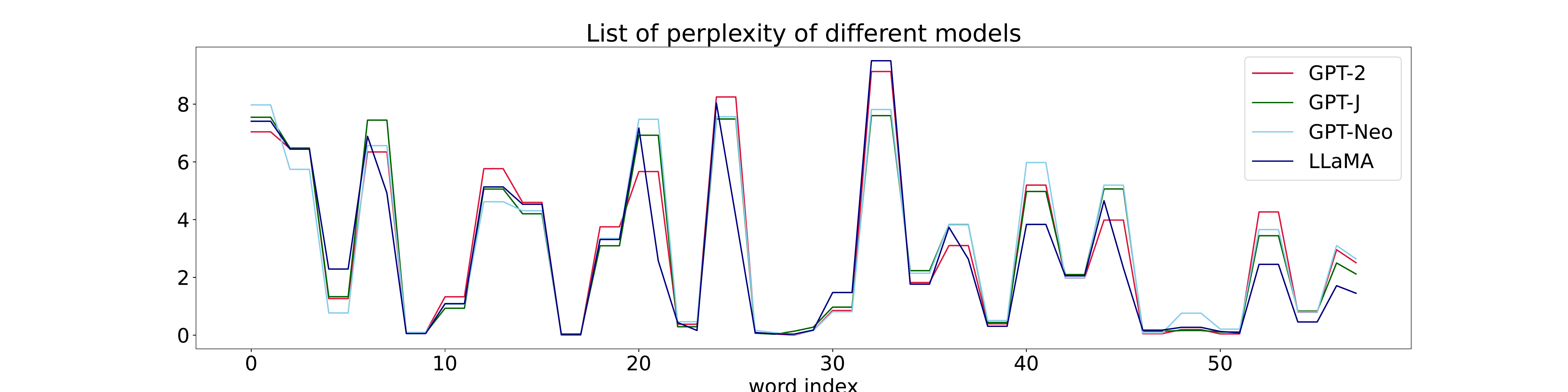} \\
    \cline{2-2}
    &  
    \multicolumn{1}{l}{
         \begin{tabular}{lll}\makecell[l]{Extracted Features: \\
           log $p(x)$ (GPT-2):  2.73 \\
           log $p(x)$ (GPT-Neo):2.77 \\
           log $p(x)$ (GPT-J):  2.72 \\
           log $p(x)$ (LLaMA):  2.19
         }
         \end{tabular}
    }\\
     & 
     \multicolumn{1}{l}{
        \begin{tabular}{lll}
           \multirow{6}{*}{pct-scores / Pearson scores / Spearman scores: }
            &  GPT-2 and GPT-Neo &  0.64/0.97/0.98   \\
            &  GPT-2 and GPT-J &    0.55/0.98/0.98   \\
            &  GPT-2 and LLaMA &    0.47/0.94/0.97   \\
            &  GPT-Neo and GPT-J &  0.38/0.99/0.98   \\
            &  GPT-Neo and LLaMA &  0.41/0.91/0.95   \\
            &  GPT-J and LLaMA &    0.41/0.93/0.97   \\
         \end{tabular}
     }
      \\
    \cline{2-2}
    & 
    \makecell[l]{Tracing Results: \\
        GPT-2: 0.8\% \\
        GPT-J/Neo: 0.3 \% \\
        LLaMA: 0.6 \% \\
        ChatGPT: \textcolor{teal}{97.3 \%} \\
        Human: 1.0 \%
     } \\
    
    \midrule
    
    \multirow{5}{*}{Human} & \makecell[l]{
       College sports are also popular in southern California. The UCLA Bruins and the USC \\ Trojans both field teams in NCAA Division I in the Pac-12 Conference, and there is a \\ longtime rivalry between the schools. \\
    } \\
    & \includegraphics[width=0.8\textwidth]{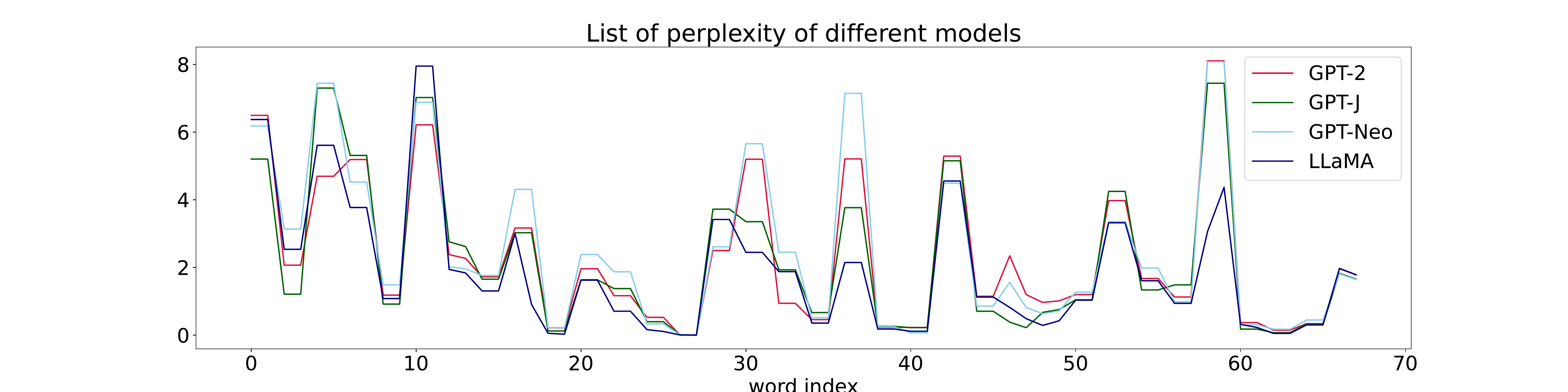} \\
    \cline{2-2}
    &  
    \multicolumn{1}{l}{
         \begin{tabular}{lll}\makecell[l]{Extracted Features: \\
           log $p(x)$ (GPT-2):  2.03 \\
           log $p(x)$ (GPT-Neo):2.21 \\
           log $p(x)$ (GPT-J):  1.95 \\
           log $p(x)$ (LLaMA):  1.74
         }
         \end{tabular}
    }\\
     & 
     \multicolumn{1}{l}{
        \begin{tabular}{lll}
           \multirow{6}{*}{pct-scores / Pearson scores / Spearman scores: }
            &  GPT-2 and GPT-Neo &  0.59/0.95/0.95   \\
            &  GPT-2 and GPT-J &    0.35/0.93/0.92   \\
            &  GPT-2 and LLaMA &    0.21/0.85/0.92   \\
            &  GPT-Neo and GPT-J &  0.41/0.93/0.95   \\
            &  GPT-Neo and LLaMA &  0.22/0.84/0.94   \\
            &  GPT-J and LLaMA &    0.26/0.91/0.94   \\
         \end{tabular}
     }
      \\
    \cline{2-2}
    & 
    \makecell[l]{Tracing Results: \\
        GPT-2: 0.0 \% \\
        GPT-J/Neo: 0.0 \% \\
        LLaMA: 14.1\% \\
        ChatGPT: 40.6 \% \\
        Human: \textcolor{teal}{45.3 \%}
     } \\

    \bottomrule
    \end{tabular}
    \caption{Random selected case studies (c).
    }
    \label{tab:casestudy-c}
\end{table}

\end{document}